# Large Language Model-Based Interpretable Machine Learning Control in Building Energy Systems

Liang Zhang[1,2]*, Zhelun Chen[3]
1. University of Arizona, Tucson, Arizona, U.S.
2. National Renewable Energy Laboratory, Golden, Colorado, U.S.
3. Drexel University, Philadelphia, Pennsylvania, U.S.
*Corresponding Author: liangzhang1@arizona.edu

## Highlights

- Innovative interpretable machine learning framework for machine learning control
- Shapley values and large language models are combined for improved interpretability
- Case study demonstrates interpretable control processes in demand response events
- Bridging trust gap in machine learning control usage for building energy management

## Abstract

The potential of Machine Learning Control (MLC) in HVAC systems is hindered by its opaque nature and inference mechanisms, which is challenging for users and modelers to fully comprehend, ultimately leading to a lack of trust in MLC-based decision-making. To address this challenge, this paper investigates and explores Interpretable Machine Learning (IML), a branch of Machine Learning (ML) that enhances transparency and understanding of models and their inferences, to improve the credibility of MLC and its industrial application in HVAC systems. Specifically, we developed an innovative framework that combines the principles of Shapley values and the in-context learning feature of Large Language Models (LLMs). While the Shapley values are instrumental in dissecting the contributions of various features in ML models, LLM provides an in-depth understanding of the non-data-driven or rule-based elements in MLC; combining them, LLM further packages these insights into a coherent, human-understandable narrative. The paper presents a case study to demonstrate the feasibility of the developed IML framework for model predictive control-based precooling under demand response events in a virtual testbed. The results indicate that the developed framework generates and explains the control signals in accordance with the rule-based rationale.

## 1 Introduction

Building operations account for approximately 30% of the world's total energy consumption and 26% of global greenhouse gas emissions (EIA 2023). As predicted by the Department of Energy [1], the US needs advanced technologies to achieve substantial decarbonization and transition to clean energy by 2050, considering the projected increases in both population and business activities. In this endeavor, the control and optimization of building operations emerge as a pivotal aspect. The precise control and adjustment of various building systems based on real-time data and conditions helps to eliminate wasteful energy consumption, prevent system conflicts, and ensure that energy-consuming equipment operates at optimal levels [2].

Given the surge in building automation systems adoption and the influx of a large amount of data, the industry's pivot to machine learning control (MLC) [3, 4] has become imperative. MLC has the capabilities to handle complex nonlinear systems where traditional linear control methods fall short [5, 6]. Machine Learning Control (MLC) applications in building energy systems have demonstrated great potential and can be categorized into different categories based on their control strategies [4]. Machine Learning-based Model Predictive Control (ML-based MPC) utilizes ML models for system identification. For instance, Cole et al. [7] employed a neural network (NN) to predict building energy load, and Kim et al. [8] used NN to forecast indoor air temperature, energy consumption, and daylight illuminance. In contrast, ML-based model-free control, such as Reinforcement Learning (RL), directly learns control actions from environmental interactions. Examples include Wei et al. [9], who applied RL for zone airflow control, and Ruelens et al. [10], who used RL for heat pump thermostat control.

Despite substantial academic explorations, the broad industrial application of MLC is still in its infancy. A key obstacle is the lack of interpretation of data-driven models due to lack of transparency and their nature as black box model [11]. Despite the high accuracy of traditional machine learning (ML) models, they are reported to lack interpretability in many fields [12]. Unlike physics-based modeling, ML models often ignore physical knowledge and information [11]. This omission introduces a level of uncertainty, complicating their reliable and transparent implementation in critical areas where interpretability and trust are paramount.

Interpretable machine learning (IML) [13] emerges as a pivotal field of research, focusing on crafting models and algorithms that not only predict but also elucidate the rationale behind their decisions and behaviors, thereby enhancing transparency and understandability for users. At its core, IML aims to bridge the gap between complex ML models and logical interpretability, ensuring that users can comprehend how and why specific predictions are made [12]. IML employs various methods such as visualizations, decision rules, and feature importance techniques to explain the behavior of the model [12, 14]. Various IML methods and tools have been developed to enhance the interpretability of ML models. **SHapley Additive Explanation (SHAP).** Proposed by Lundberg and Lee [15] in 2017. The SHAP values are based on the concept of Shapley values from cooperative game theory, which measures the "fair" contribution of each feature to the prediction outcome [12] [13]. These SHAP values represent the expected change in the prediction when a particular feature is included or excluded from the model. **Local Interpretable Model-agnostic Explanations (LIME).** Introduced by Ribeiro [6] in 2016, LIME is a tool to provide interpretable explanations for individual predictions made by complex ML models. LIME works by approximating the behavior of a ML model locally using a simpler interpretable surrogate model, such as a linear model or a decision tree. This surrogate model then helps in extracting explanations about the decision-making process of the original model [6]. **Shapash.** Introduced by Saabas and Parikh in 2020, Shapash [16] is a Python library that combines the interpretability abilities of SHAP or LIME and the interactive visualization capabilities of Dash. This allows users to explore and understand the predictions made by ML models in a more interactive manner. **Machine Learning Extensions (MLxtend).** MLxtend is a library that implements a variety of core algorithms and utilities for ML and data mining [17]. The primary goal of MLxtend is to make commonly used tools accessible to researchers in academia and data scientists in industries focusing on user friendly and intuitive application programming interfaces (APIs) and compatibility to existing ML libraries, such as scikit-learn, when appropriate.

IML has been studied across different research topics in building operations. In the field of **energy modeling**, Li. et al. [18] introduced a novel neural network architecture, incorporating an attention mechanism into Recurrent Neural Networks to enhance interpretability in building energy predictions. Chang et al. [19] propose a unified clustering-based prediction framework to provide short-term prediction of PV power output with interpretability analysis to reveal the features that are important for the prediction. Papadopoulos and Kontokosta [20] developed a building energy performance grading methodology using ML and city-specific energy use and building data. They applied IML tool SHAP in green grading system to better benchmark green buildings. Fan et.al. [21] proposed a comprehensive methodology based on IML method LIME to explain and evaluate data-driven building energy performance models. In the field of **occupant comfort modeling**, Jin et al. [22] applied IML tool SHAP to assess interpretability for occupant preferences by ranking the non-dominated solutions using multi-criteria decision analysis and selecting a subset of consistent and reasonable operating strategies. Similarly, Yang et al. [23] performed an interpretability analysis using SHAP to explain black-box thermal sensation models. The effects of different features on thermal sensations and typical decision routes in the models were investigated from both local and global perspectives, and the properties of correlation between features and thermal sensations and decision routes within ML models were summarized. Bellahsen and Dagdougui [24] developed a comparative study for aggregated short-term load forecasting, which applied SHAP to determine the contribution of features to the peak load. Ugwuanyi [25] applied IML method SHAP and LIME for indoor $CO_2$ level prediction and occupancy estimation. Zhang [26] proposed an interpretable thermal comfort system to introduce interpretability to any black-box comfort models. They focus on 1) the relationship between a model's input features and output comfort level and 2) the model mechanisms about the data processing inside the model by building the model surrogates based on IML. In the field of **building control**, which is the main **focus of this paper**, Yu et al. [27] extracted simple and interpretable supervisory building control rules from multiple optimal solutions that combined multi-criteria decision analysis, control strategy clustering, and classification tree models. They applied low-dimensional classification trees for high interpretability and ease of implementation. Kotevska et al. [28] applied interpretability tools and systematically exploring the model inputs for better understanding the Deep Reinforcement Learning control and decision-making process. The interpretability tools and methods include a model agnostic interpretability technique [29], individual conditional expectations [30], and partial dependence plot [30]. Park et al. [31] utilized SHAP for generating local and global interpretations for the differential pressure control valve to detect anomalies in a district heating system. Mao [32] explored the use of an IML technique called SHAP to create models for model predictive control (MPC) for an HVAC unit. A new IML algorithm called Permutation Feature-based Frequency Response Analysis was also proposed. Gao [33] utilized the SHAP technique to explore black-box system-level chiller faults.

While there is clear applicability across various domains reviewed earlier, most of the research focuses on applying IML in quantifying and ranking feature importance, still relying on modelers' expert knowledge for deeper insights into "what these important features reveal". Understandings of these results vary among authors and across different applications, necessitating domain expertise and tailored analysis by IML tool users. There remains a significant gap as most current research falls short of fully utilizing IML techniques for in-depth analysis and comprehensive exploration of these findings in the context of specific building modeling and control tasks, leaving a critical gap in the complete and automated interpretation and understanding of the results.

Large language models (LLMs), such as GPT [34] and Llama [35], emerge as a promising avenue to address this challenge. LLMs are characterized by their ability to comprehend and generate text in a nuanced manner, understand context, and provide detailed responses or analyses. They can be employed to undertake a deeper analysis of the results generated through IML studies in building control. By utilizing IML tools like SHAP alongside LLMs, researchers can dissect the complex patterns and relationships in the data more effectively. LLMs can enrich the understanding of how different inputs influence specific control tasks, leading to a deeper, more nuanced analysis. Furthermore, they have the inherent capacity to synthesize information from a multitude of sources to create coherent and logically structured narratives. This characteristic stands them in good stead to address the challenges raised in the systematic arrangement of interpretations derived from multiple ML predictions in building control tasks. LLMs can be designed to aggregate and analyze the diverse sets of predictions, creating a "storyline" that logically connects individual predictive elements to offer a holistic view of the multi-step decision-making process. By doing so, they can construct a rational pathway that delineates the series of control actions necessary based on the predictions. The integration of IML and LLM would significantly enhance the transparency and effectiveness of IML strategies.

This study develops an innovative framework that seamlessly combines the principles of Shapley values from game theory and the capabilities of LLMs to enhance the interpretability of MLC in building energy management. Through the illustrative case study, the framework showcases its potential to bridge the trust gap between algorithm developers and building operators, ensuring successful adoptions of advanced AI controls in real-world settings. Although this study discusses the broader spectrum of MLC applications, including references to both ML-based MPC and RL control, it is important to note that our primary focus of the case study centers on ML-based MPC. Throughout the remainder of this paper, the term 'MLC' will specifically denote ML-based MPC unless otherwise specified.

The content of this paper is organized as follows. We will first introduce the methodology of the developed framework, which integrates Shapley values with LLM to explain step-by-step predictions and LLMs to streamline the interpretation of the whole picture of ML-based MPC. Next, we will implement the proposed framework in a virtual building case study based on DOE reference small office building [36]. Moreover, we will analyze the results of the case studies and discuss the potential impact of this framework on building energy management. The final section presents the conclusions, limitations, and future work.

# 2 Methodology

We introduce a novel IML framework for MLC in buildings, integrating Shapley value theory and LLM. This framework offers detailed reasoning and explanations for the ML subprocesses involved in the MLC.

## 2.1 SHapley Additive exPlanation (SHAP)

SHAP is a method in cooperative game theory that assigns a unique distribution of the total gain of a coalition to its players, based on their marginal contribution. It was named after Lloyd Shapley, who introduced it in his paper in 1953 [37]. In the context of IML, it has been adapted to measure the importance of features in prediction models. Formally, given a game with n players, let N={1,2,...,n} be the set of players, and let $v:2^N \rightarrow R$ be a characteristic function that assigns a real number (the worth) to each subset (coalition) of N. The Shapley value of player i in the game (N,v) is given by:

$$\phi_i(N, v) = \sum_{S \subseteq N \setminus \{i\}} \frac{|S|!(n-|S|-1)!}{n!} [v(S \cup \{i\}) - v(S)] \qquad \text{Equation 1}$$

Where S ranges over all subsets of NN that do not include player i, $|S|$ denotes the cardinality of set S, and n is the total number of players. The term $\frac{|S|!(n-|S|-1)!}{n!}$ is the weight assigned to each coalition, representing the number of permutations of N where player i joins the coalition S as the $(|S|+1)^{th}$ member, divided by the total number of permutations of N.

In Equation 1, the term $v(S \cup \{i\}) - v(S)$ represents the marginal contribution of player i to coalition S. The Shapley value is then calculated as the weighted sum of these marginal contributions over all possible coalitions that player i can join.

In the context of IML, each "player" is a feature of the model, and the "coalition" is a subset of features. The "game" is the prediction task, and the "worth" of a coalition is the prediction performance of the model when it is trained using only the features in the coalition. The Shapley value of a feature is a measure of its average contribution to the prediction performance, considering all possible combinations of features. This is particularly important when trying to understand the decision-making process of complex models, where individual feature contributions can be obfuscated by the complexity of the model's architecture. The Shapley values framework thus offers a robust, theoretically-grounded approach to calculating the relative importance of each input feature to the model's final prediction.

The output of Shapley consists of variables, their corresponding values, and their contributions to the output, as can be seen from an example SHAP explanation shown in **Figure 1**. Conventionally, ML models are black-box without knowing the contribution of features to a specific inference. With SHAP, the contributions of variables are quantified, which can benefit the interpretation of the output. In this case, we know that "age" has the largest positive Shapley value (+0.4), indicating the largest positive impact on the output; "sex" has the largest negative Shapley value (-0.3), indicating the largest negative impact on the output. The sum of Shapley values of all features (+0.4 − 0.3 + 0.1 + 0.1 = 0.3) plus the base rate (0.1) equals the inference value or output of the model (0.4).

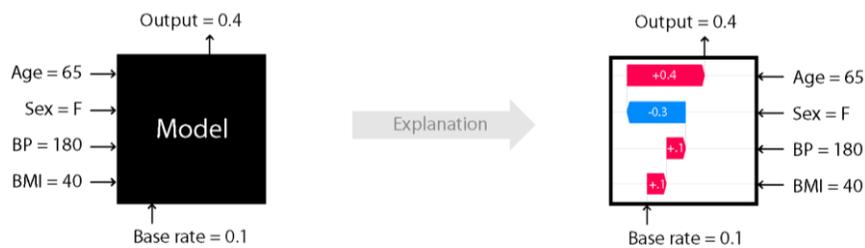

**Figure 1**. SHAP explanations example (source: [38])

IML techniques, such as SHAP, excel in generating inferences from single models. However, their application becomes convoluted in the face of complex engineering problems, such as MLC which involves multiple black-box models and intricate control formations. IML techniques tend to provide a limited view, offering insights only into partial or local aspects of the problem. For instance, in building control scenarios, a control decision is often influenced by predictions from multiple models. Understanding the collective impact of these predictions on the final decision can be challenging. Moreover, while IML techniques provide feature importance, the deeper rationale behind these features often remains elusive, reducing their self-explanatory nature. Consequently, there is a need for a systematic IML framework specifically designed for MLC.

## 2.2 IML Framework Combining LLM for MPC

In this paper, we introduce a novel IML framework for MLC applications, integrating Shapley value theory and LLM. This framework offers detailed reasoning and explanations for the black-box, data-driven, and machine-learning subprocesses involved in the MLC.

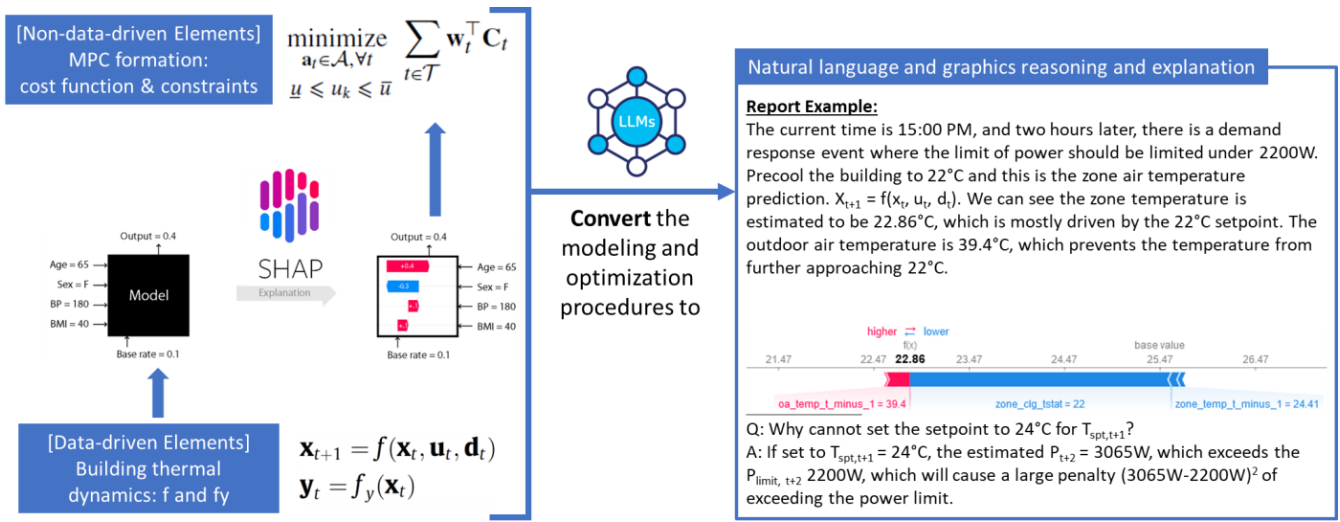

**Figure 2.** Diagram of the developed IML framework in MPC

In the context of MPC, the explanation process begins with the current status of state variables, inputs, and outputs of the MPC. Subsequently, the IML framework elucidates predictions, optimizations, and actions in a structured, step-by-step manner, based on control formulations such as cost functions, constraints, and state space models. The proposed framework generates explanatory text and graphs systematically, enhancing interpretability. In this framework, LLM serves as a post-processing and integration tool, summarizing explanatory text and graphs from IML tool SHAP. Figure 2 presents a schematic representation of the proposed IML framework under the ML-based MPC architecture. The foundational steps in this IML framework involve:

1) Step 1. Using an LLM to understand the non-data-driven elements within the formulation of MPC. The formulation of MPC includes non-data-driven cost functions and constraints, and data-driven state space models. We will feed this information to LLM as a prompt. The outcome of this step is an LLM model that is aware of the MPC setting and goals.

2) Step 2. Using SHAP to generate interpretability results of each ML model describing thermal dynamics of MPC (e.g., predicting temperature, thermal comfort, weather, and energy consumption) and using LLM to enhance the interpretability of the generated Shapley values. **Figure 3** is an example of SHAP results from the prediction of cooling load of a room in the next hour. Although it shows quantitative Shapley values of the contribution of the results, further descriptive analysis in the cooling load prediction and MLC context is missing. The following text is a good example of what is missing:

"*According to the Shapley value analysis, the three most impactful features on cooling energy consumption are oa_temp, zone_temp, and zone_occ. The oa_temp, or outdoor air temperature, has a negative impact on cooling energy consumption, suggesting that higher outdoor temperatures require more cooling energy. Similarly, the zone_temp, or zone temperature, also has a negative impact, indicating that higher indoor temperatures result in increased cooling energy consumption. On the other hand, the zone_occ, or zone occupancy, has a positive impact, implying that higher occupancy levels lead to higher cooling energy consumption. These findings align with common sense, as hotter outdoor and indoor temperatures and increased occupancy typically require more cooling to maintain a comfortable environment. The expected value of 1484.28 represents the average cooling energy consumption based on these impactful features.*"

And this description is exactly what LLM generates based on the Shapley value, along with its integrated building physics knowledge. The LLM's aptitude for interpreting Shapley values and producing such explanations highlights its effectiveness. This capability arises from its ability in the reasoning from general physics (building physics in this case).

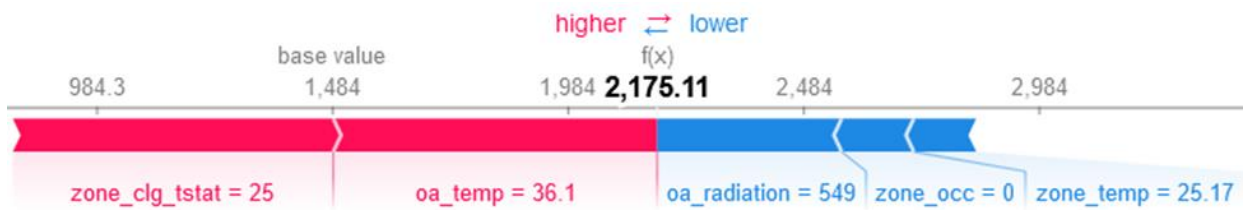

**Figure 3**. Example of SHAP results for interpreting the prediction of next hour room temperature

3) Step 3. Consolidating the non-data-driven components in step 1 and the ML-based elements, as interpreted by SHAP in step 2, into a cohesive narrative as shown on the right part of Figure 2. This process results in the generation of a comprehensive text-graph document that details the rationale behind the MLC decision process. This document serves to inform stakeholders about the sequence of events and the logic that underpins the decisions within the MLC, thereby enhancing interpretability and transparency. Overall, the document includes: 1) all data information in each timestep, including time, state variables, disturbances, controllables, SHAP results, and any other sensor data or information fed into LLMs per timestep, 2) predictions from all ML models in MPC, along with SHAP-based rationale represented through Shapley values and graphs (e.g., **Figure 3**), for each timestep, 3) the rationale behind decision making and optimization results from MPC per timestep. An example rationale is that MPC decides to precool the building because of the anticipation of a demand response event to avoid penalties. The document can be tailored based on user's specific need of arranging the above content, and this is defined by the prompt template which will be introduced in the next paragraph. The detail level of the document can be also tailored to the user's requirements, leveraging the LLM's ability to summarize and elaborate content: the example on the right part of Figure 2 presents a concise version; a more detailed example is available in the case study section.

Specifically, prompt engineering techniques are applied as the key LLM technique to develop a module/agent to 1) organize and generate documentation, and 2) explain and interpret MLC. Prompt engineering refers to the technology of crafting specific and targeted queries or statements that guide LLMs, to generate desired and contextually relevant outputs. By structuring and refining these prompts, we can extract more precise and interpretable insights from the model, ensuring that the explanations are both detailed and aligned with the objectives of the IML framework. Figure 4 describes the developed prompt engineering framework. The framework starts from getting MLC variables, including time, state variables, disturbances, controllables, and any other sensor data or information fed into LLMs per timestep. Then, the Shapley values with variable values are combined as the source data to generate descriptive sentences to interpret ML predictions. Prompt template serves as a foundational framework, aggregating all the components into an organized storyline. Then the Complete Prompt is fed into LLM to generate MLC explanation document and enable Q&A function. We will detail this process in the case study.

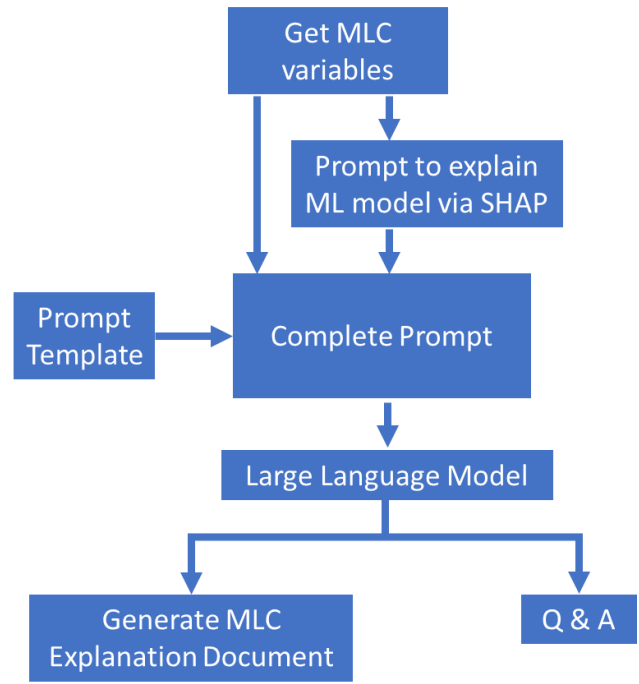

**Figure 4.** Detailed prompt engineering framework for IML

## 3 Case Study

A case study is designed to demonstrate the effectiveness of the developed IML framework. The case study is conducted in a virtual building co-simulated with MLC. The MLC precools the building to respond to demand response event. We will use the developed IML framework to generate summarized explanations and Q&A module for users to understand and interact with the control decision from the MLC.

### 3.1 Virtual Building

The object building in the case study is the small reference office building developed by National Renewable Energy Laboratory [36] under the environment of EnergyPlus [39]. It is a one-story 511-square-meter office building; the packaged air-conditioning unit is used for cooling and furnace is used for heating; the air distribution system is single-zone constant air volume; floor-to-ceiling height is 3.05 meter; the window-to-wall ratio is 0.21 meter; lighting follows the code of ASHRAE 90.1-2004. Figure 5 shows the outlook and zone configuration of the building. More details of the small office building can be found in [36]. The building model is simulated under the TMY (typical meteorological year) weather file "USA_AZ_Tucson.Intl.AP.722740_TMY3.epw". The MLC is conducted in one zone of the building (Perimeter_ZN_1) for simplified demonstration purposes.

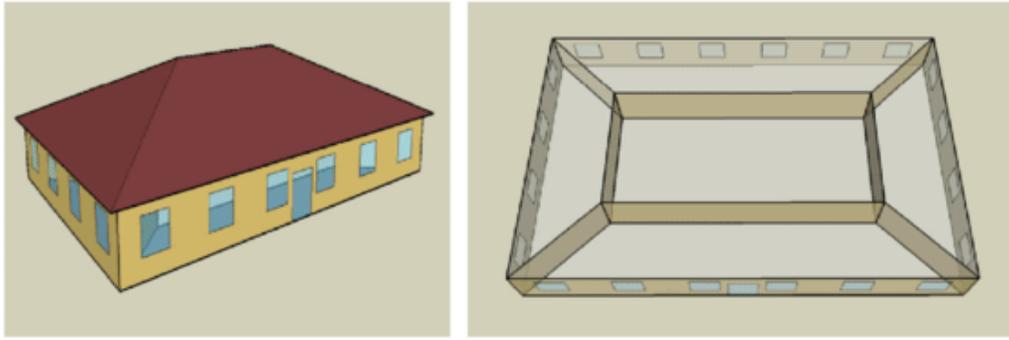

**Figure 5.** The appearance and zone configuration of the small office building

The run period of ML model training of building dynamics with random excitation is July 1st to 31st; the run period of testing MPC is August 1st to 31st. Supervisory control is implemented by adjusting the cooling setpoints. And the constraints of the cooling setpoint is from 22 to 26°C. The heating setpoint is set to 20°C; the run period is in summer, so no heating is observed. The HVAC schedule is always on regardless of occupancy. All the schedules are in their default value in the reference building model.

### 3.2 Py-CoSim

We use a co-simulation testbed called Py-CoSim to connect the MLC and the virtual building. Py-CoSim platform [40] is a Python-EMS-based building energy co-simulation platform. The platform enables co-simulation among multiple buildings, on-site power generation, energy storage, grid operations, advanced building control, transportation system, decision-making, optimization, etc. all under Python environment. In this modeling platform, each co-simulation component is represented by a formatted Python class or function. The communication among components is realized by Simulation Management and Messaging Module, which serves as a data hub that saves, reads, and manages real-time energy simulation results, and synchronizes multiple building runs. The structural diagram of Py-CoSim is shown in Figure 6. All data for machine learning modeling is generated from the virtual EnergyPlus model (introduced in Section 3.1) running under Py-CoSim testbed.

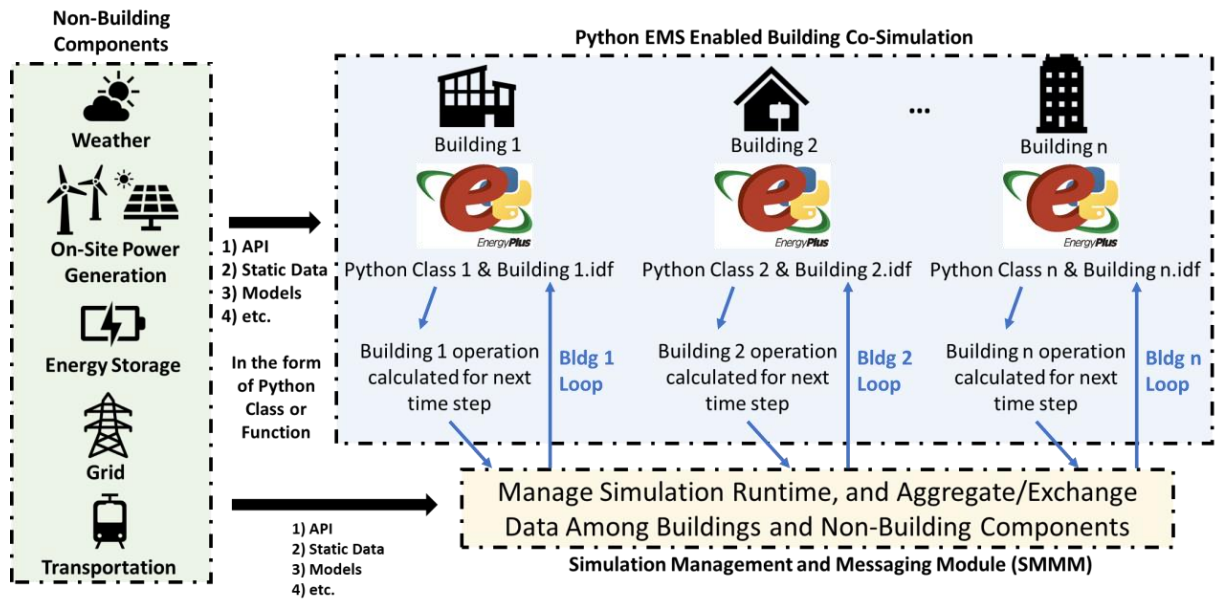

**Figure 6.** Structural Diagram of Py-CoSim

The MLC controller is defined and modeled via a Python Function, and the building is defined and modeled via a Python Class based on Python EMS under Py-CoSim co-simulation platform.

### 3.3 MLC Settings

The MLC in the case study is in essence a ML-based MPC. The controllable input (u) is zone temperature setpoint (ZTSP); the state variable (x) is zone air temperature (ZSF1_ZT); the disturbances (d) include outdoor air dry-bulb temperature (OADT), direct solar radiation rate per area (SR_DIR), and occupancy (OCC); the output (y) is single zone cooling rate (Clg_Rate). The MLC formation is defined in Equation 2-6.

$$\min_{u1,u2} \sum_{k=1}^{2}(y_k + V_k)$$

Equation 2

$$s.t. \ \ y_k = f_y(u_k, x_k, d_k)$$

Equation 3

$$x_k = f_x(u_k, x_{k-1}, d_{k-1})$$

Equation 4

$$u_1, u_2 \ \in [22, 26]°C \ ([71.6, 78.8]°F)$$

Equation 5

$$V_k = \begin{cases} (P - P_{limit})^2, P > P_{limit} \\ 0, else \end{cases}$$

Equation 6

As can be seen from the cost function defined in Equation 2, the forecasting horizon is 2 hours ($u1, u2$ are the first- and second-hour cooling setpoint, respectively). In each timestep, the MLC will generate two optimal cooling setpoints for the next hour and the hour after the next. The cost function is defined by two parts: cooling rate $y_k$ and the penalty of violating the demand response event $V_k$. Cooling rate $y_k$ is calculated by two ML models. In Equation 3, $x_k$ is first calculated with the ML model for zone temperature prediction $f_x$. With the calculated $x_k$, based on Equation 4, the cooling rate is calculated with the ML model for energy prediction denoted as $f_y$. The ML algorithms used for $f_x$ and $f_y$ are deep learning algorithm named deep feed forward neural net model developed in PyTorch. It is a type of deep learning algorithm that processes information through sequential layers, transforming inputs into outputs without recurrent connections. In terms of the hyperparameters, the hidden dimension is 50, number of epochs is 10000, loss function is MSE, and the optimizer is based on Adam algorithm. Based on the small number of inputs of the model, we do not conduct feature selection. In order to get unbiased training data to accurately model $f_x$ and $f_y$, we conduct random excitation of zone temperature setpoint to randomly change it from 22 to 26°C with 1 hour interval lasting 1 month simulation time (July 1st to 31st). Equation 5 imposes constraints on the range of setpoints to guarantee thermal comfort within the zone. In Equation 6, as the second part of the cost function, power limit penalty $V_k$ is a piece-wise function: when the cooling rate is greater than power limit, $V_k$ is calculated by the squared difference of the exceeding power; otherwise, $V_k$ is 0 meaning there is no penalty if the cooling rate does not exceed power limit. Power limit $P_{limit}$ indicates the demand response event. In normal situations, $P_{limit}$ = 5000W (6.7hp), which is a high bar that normally will not be exceeded in the virtual testbed. In demand response event, $P_{limit} \sim$ U(750W(1.0hp), 1750W(2.3hp)), meaning that it follows a uniform distribution with maximum value of 1750W (2.3hp) and minimum value of 750W (1.0hp). Also, demand response happens randomly from 11:00am to 18:00 pm. The duration of the demand response event is 1 hour. The exhaustive searching method (or brute force method) is used as the optimization method in this study due to a small variable space. The exhaustive searching method first discretizes the set-points by 1K (1.8R) from 22 to 26°C (71.6 to 78.8°F) throughout the control horizon. Then it calculates the cost function of every combination of setpoints throughout the control horizon and finds the optimal control sequence.

### 3.4 LLM Settings

In this case study, OpenAI's premier LLM, ChatGPT, was utilized via its API. The specific model version employed
was "gpt-3.5-turbo-0301." The temperature setting of a Large Language Model (LLM), a parameter controlling
the balance between predictability and creativity in generated responses, is set at 0.5 in the case study. This
setting strikes an optimal balance, ensuring responses are both relevant and appropriately varied.

# 4 Results and Discussion

## 4.1 Results: MPC

The MLC results, along with the interpretable parts of MLC results of the case study, are presented and analyzed
in this section. Figure 7 displays the results of the deep learning MPC applied for precooling buildings in response
to demand response events. The figure shows that 1) precooling does not occur for every demand response
event, and 2) the precooling setpoints fluctuate based on the power limit. Like many deep learning controls, the
outcomes are not inherently interpretable, potentially hindering building operators' complete trust in the
decisions made by the MPC. We conducted the MPC simulations on personal computers configured with an 11th
Gen Intel(R) Core(TM) i7-1185G7@3.00GHz, 1.80GHz processor and 16GB RAM. The building co-simulation
period spanned 31 days, from August 1st to 31st, with a 1-hour interval, taking a total of 3120 seconds to
optimize objective function for MPC in 31 days * 24 hours/day = 744 steps —equivalent to 4.19 seconds per
control interval. Despite the inherent slowness of the brute force algorithm, the decision pace remains relatively
quick and within an acceptable range.

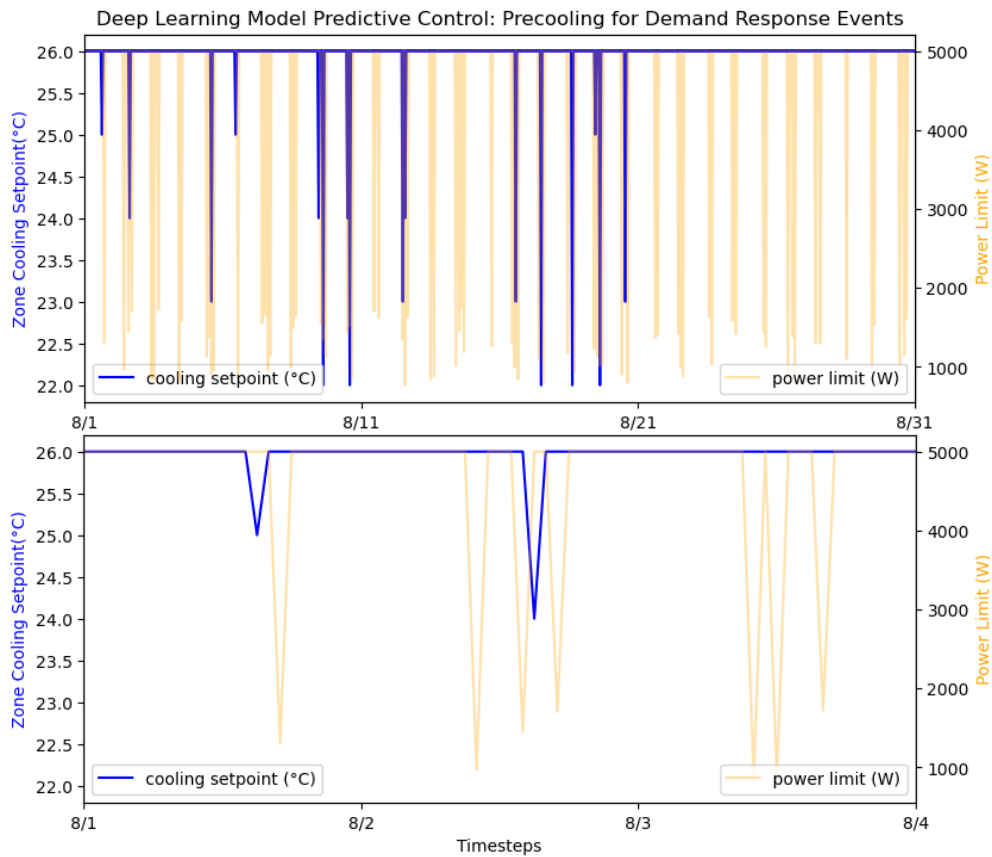

**Figure 7.** Deep learning MPC results: precooling for demand response events

## 4.2 Results: IML Framework Combining LLM for MPC

We followed the procedures in the proposed framework (Figure 4).

**Step 1. Get Py-CoSim MLC variables.** The framework starts from getting variables from MLC variables from Py-CoSim. The variables include 1) future power limits, 2) controllables, state variables, disturbances, and output at the current timestep, 3) SHAP explanation results, and 4) MLC results, which are temperature setpoints in the next hour and the in the hour after the next.

**Step 2. Prompt to explain ML model via SHAP.** Then, the Shapley values with variable values are combined as the source data to generate descriptive sentences to interpret ML predictions. The prompt engineered to explain SHAP results is shown below.

---

**Prompt:**

Based on the Shapley value and variable values, please help me generate a descriptive paragraph:

Shapley values: oa_temp 680.369781; oa_radiation 33.052102; zone_temp 18.838554; zone_clg_tstat -113.826475; zone_occ -98.523013; expected_value 1544.673602

Variable values: oa_radiation 332; oa_temp 37.2; zone_clg_stat 26; zone_occ 0

Only introduce the largest 3 impactful features plus the expected value. Also, explain the potential reason why these features are impactful to the cooling energy consumption

The variable dictionary is listed as follows: {zone_clg_stat: zone temperature setpoint, zone_temp: zone air temperature, oa_temp: outdoor air dry-bulb temperature, oa_radiation: direct solar radiation rate per area, zone_occ: occupancy}

---

And the response from LLM is shown as follows. The LLM's aptitude for interpreting Shapley values and producing such explanations highlights its effectiveness. This capability arises from its ability in the reasoning from general physics (building physics in this case).

---

**Response:**

The Shapley values indicate the contribution of each feature to the prediction of the cooling energy consumption. The three most impactful features are the outdoor air dry-bulb temperature (oa_temp), the zone temperature setpoint (zone_clg_tstat), and the occupancy (zone_occ). The outdoor air dry-bulb temperature has the highest Shapley value of 680.37, suggesting it has the most significant impact on the cooling energy consumption. This is in line with common sense as the outdoor temperature directly influences the indoor temperature and thus the amount of energy required to cool the building. The higher the outdoor temperature, the more energy is needed for cooling. The zone temperature setpoint has a Shapley value of -113.83, indicating that it negatively impacts the cooling energy consumption. This makes sense as the higher the setpoint temperature, the less the cooling system needs to work, resulting in lower energy consumption. The occupancy has a Shapley value of -98.52, also indicating a negative impact on the cooling energy consumption. This is logical as the more people in the building, the more heat is generated, leading to an increase in the cooling energy consumption. However, in this case, the occupancy is zero, which means no additional heat is generated by people, thus reducing the cooling energy consumption. The expected value of the cooling energy consumption is 1544.67. This value is the prediction made by the model when considering all the features and their interactions.

---

**Step 3. Prompt template.** The template defines three different scenarios particularly for this MLC case study: 1) predicted demand response event and pre-cool the building, 2) no demand response event and keep the building operate normally, 3) predicted demand response event, but the building is not pre-cooled. A segment of the prompt template is presented below.

---

**Prompt Template:**

Scenario 1: The criteria of Scenario 1 is Plimit(t+2) < Plimit(threshold) and Tspt(t+1) < 26°C. Generate a paragraph like the following: "Scenario 1 criteria: Since Plimit(t+2) = 2500W < 5000W, meaning there is a demand response event in the hour after the next. The MPC calculated that the building should be first pre-cooled to Tspt (t+1) = **[placeholder]**°C, which is less than 26°C. According to the ML zone air temperature prediction Xt+1 = fx(xt, ut, dt), the zone temperature is estimated to be Tz(t+1) = **[placeholder]**°C, which can be explained by the SHAP plot shown as follows **[figure 1 placeholder] [model 1 SHAP explanation placeholder].** Based on the predicted temperature of Tz(t+1), we can further use ML model yt+1 = fy(xt+1, ut+1, dt+1) to predict the cooling power rate in the next hour Pt+1 = **[placeholder]**W, which can be explained by the SHAP plot shown as follows **[figure 2 placeholder] [model 2 SHAP explanation placeholder].** In the second forecasting horizon, MPC calculated the setpoint to Tspt(t+2) = **[placeholder]**, the second hour's temperature will be Tz(t+2) = **[placeholder]**, which can be explained by the SHAP plot shown as follows **[figure 3 placeholder] [model 3 SHAP explanation placeholder].** The corresponding cooling power is P(t+2) = **[placeholder]**W, which can be explained by the SHAP plot shown as follows **[figure 4 placeholder] [model 1 SHAP explanation placeholder].** The cooling power in the second hour P(t+2) = **[placeholder]**W is less than Plimit(t+2) = 2500W and avoided the penalty from the demand response event."

Scenario 2: …. Scenario 3: … [due to word constraints, this section has been omitted]

Based on the following inputs, judge what kind of scenario it is, and then generate the corresponding paragraph:**[Placeholder for Py-CoSim MLC variables]**

---

**Step 4. Generate Explanation Document from Complete Prompt.** Based on the prompt template (Step 3), we fill the placeholder in the template with the collected data from Py-CoSim MLC variables (Step 1), SHAP plots, and Prompt to explain ML model via SHAP (Step 2). We then feed the complete string to LLM so that LLM can automatically identify the control scenario of the MPC (mentioned in Step 3). The subsequent paragraphs present final output of the framework, which is the **explanation document**, derived from the comprehensive prompt response provided by the LLM based on the complete prompt we input.

---

**Response:**

In timestep 8/1 15:00PM: **Future power limits**: Plimit(t+1) = 5000.0W, Plimit(t+2) = 1311.2W, Plimit(threshold) = 5000W. **Controllables, state variables, disturbances, and output:** Tz(t) = 26.0°C, OAT(t) = 36.1°C, Radiation_direct(t) = 549.0W/m$^2$, OCC(t) = 0.0, Tspt(t+1) = 25.0°C, Tspt(t+2) = 26.0°C, Tz(t+1) = 25.2°C, P(t+1) = 2175.1W, Tz(t+2) = 25.6°C, P(t+2) = 829.5W. MPC results: Tspt(t+1) = 25.0°C, Tspt(t+2) = 26.0°C.

Since Plimit(t+2) = 1311.2W < Plimit(threshold) = 5000W, there is a demand response event in the hour after the next. The MPC calculated that the building should be first pre-cooled to Tspt(t+1) = 25.0°C, which is less than 26.0°C.

---

To interpret the MLC results: According to the ML zone air temperature prediction Xt+1 = fx(xt, ut, dt), the zone temperature is estimated to be Tz(t+1) = 25.2°C if Tspt(t+1) = 25.0°C, which can be explained by the SHAP plot shown as follows.

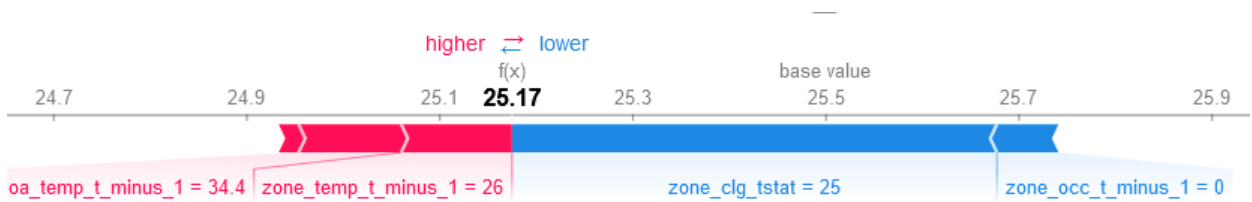

The three most impactful features on the predicted zone temperature, according to the Shapley values, are the outdoor air temperature at time t-1 (oa_temp_t_minus_1), the zone temperature at time t-1 (zone_temp_t_minus_1), and the zone temperature setpoint (zone_clg_tstat). The outdoor air temperature at time t-1 has a Shapley value of 0.89, indicating a strong positive impact on the predicted zone temperature. This is in line with common sense as the outdoor temperature is a significant factor in determining the indoor temperature, especially if the building's insulation is not very effective. The actual value of this feature is 34.4, which is relatively high and would likely contribute to a higher indoor temperature. The zone temperature at time t-1 has a Shapley value of 0.98, suggesting an even stronger positive impact on the predicted zone temperature. This makes sense as the temperature in the zone at a previous time point would naturally have a strong influence on the temperature at the current time point. The actual value of this feature is 26, which is also relatively high and would contribute to a higher current temperature. The zone temperature setpoint has a Shapley value of -0.5, indicating a negative impact on the predicted zone temperature. This is logical as the setpoint is the temperature that the air conditioning system is trying to achieve. If the setpoint is lower, the system would work harder to cool the zone, resulting in a lower predicted temperature. The actual value of this feature is 25, which is lower than the previous zone temperature, hence the negative impact. The expected value is 25.5, which is the average outcome if we had no information about the other features. This value serves as a baseline to compare the impacts of the individual features.

Based on the predicted temperature of Tz(t+1), we can further use the ML model yt+1 = fy(xt+1, ut+1, dt+1) to predict the cooling power rate in the next hour P(t+1) = 2175.1W, which can be explained by the SHAP plot shown as follows.

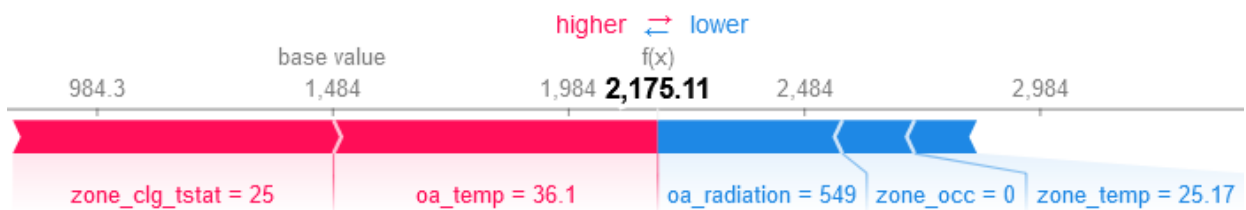

The three most impactful features on the cooling energy consumption, according to the Shapley values, are the outdoor air dry-bulb temperature (oa_temp), the zone temperature setpoint (zone_clg_tstat), and the direct solar radiation rate per area (oa_radiation). The outdoor air dry-bulb temperature has a Shapley value of 691.11, indicating a significant impact on the cooling energy consumption. This is in line with common sense as the outdoor temperature directly influences the indoor temperature and thus the amount of energy required to cool the building. The higher the outdoor temperature, the more energy is needed for cooling. The zone temperature setpoint also has a substantial impact on the cooling energy consumption, with a Shapley value of 692.23. This is logical as the setpoint determines the desired indoor temperature. If the setpoint is low, more energy is required to achieve and maintain this temperature, especially if the outdoor temperature is high. The direct solar radiation rate per area, however, has a negative Shapley value of -301.88. This suggests that an increase in solar radiation decreases the cooling energy consumption, which might seem counterintuitive at first. However, this could be due to the fact that solar radiation is often used to generate renewable energy, which could offset the energy required for cooling. The expected value, which is the average prediction for the cooling energy consumption, is 1484. This value is influenced by the aforementioned features and their respective values: an outdoor air dry-bulb temperature of 36.1, a zone temperature setpoint of 25, and a direct solar radiation rate per area of 549.

In the second forecasting horizon, MPC calculated the setpoint to Tspt(t+2) = 26.0°C, and the second hour's temperature will be Tz(t+2) = 25.6°C, which can be explained by the SHAP plot shown as follows.

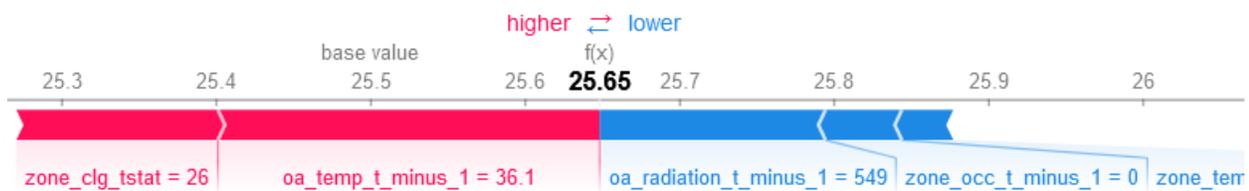

Based on the Shapley value analysis, [due to word constraints, this section has been omitted].

The corresponding cooling power is P(t+2) = 829.5W, which can be explained by the SHAP plot shown as follows

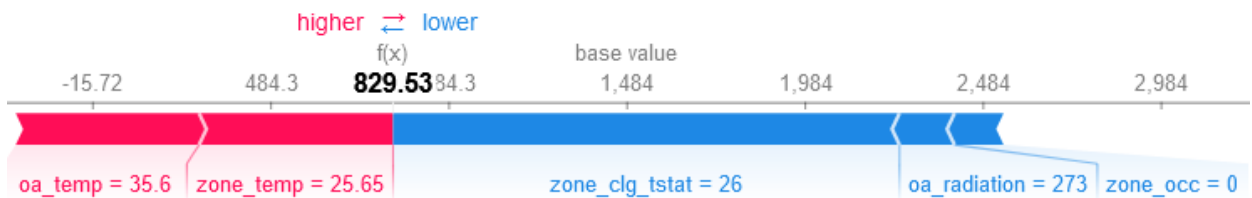

Based on the Shapley value analysis, [due to word constraints, this section has been omitted].

The cooling power in the second hour P(t+2) = 829.5W is less than Plimit(t+2) = 1311.2W, which avoids the penalty from the demand response event.

Besides the detailed document on how the decision is made, with the powerful support of LLM, if the building operator has more questions, they can ask questions based on the document as well as the strong ability of reasoning and learning ability of LLM to learn building physics and building control. An example question and answer are shown as follows.

***Example Question***:

| Prompt: |
| --- |
| I know it is suggested that precooling temperature setpoint should be set to 25°C. What will happen if I ignore the demand response signal and keep the setpoint to 26°C? |

***Example Answer***:

---

**Response:**

If the precooling temperature setpoint is set to 26°C (78.8°F) (not precooled), it means that the building will not be pre-cooled before the demand response event. This decision ignores the demand response event and does not take any action to reduce the cooling load in the second hour.

Without precooling, the cooling power in the second hour may be higher than necessary, resulting in increased energy consumption and potentially exceeding the power limit. This can lead to penalties or additional costs associated with exceeding the power limit. Therefore, not precooling the building and setting the temperature setpoint to 26°C (78.8°F) may not be an optimal strategy in this scenario, as it ignores the demand response event and can result in higher energy consumption, discomfort, and potential penalties.

---

It can be seen from the answer that the developed Q&A module successfully answers the what-if question based on the summary document as well as the building physics and control knowledge. Users can ask more questions or new ones as needed. This helps people better understand MLC by addressing their specific concerns.

To validate the scenario identification success across all 744 timesteps from August 1st to August 31st, we observed a 100% success rate (Scenario 1: 16/16, Scenario 2: 654/654, Scenario 3: 74/74) using our proposed IML framework. This high success rate is attributed to 1) straightforward judgement criteria based on our prompt templates, 2) the limited number of predictable scenarios, and 3) the robust performance of the GPT model in structured tasks. For the validation of SHAP values and Q&A accuracy, we conducted targeted evaluations during the first 3 days of the test period (August 1st to 3rd). Within this timeframe, the explanations and answers regarding why certain timesteps were classified under specific scenarios were correct. Although this testing period is brief, the results within this window demonstrate the model's consistency and reliability in its outputs.

## 4.3 Potential impact of this framework

Building operators, often from facility management, are typically not the developers of the building's control algorithms, especially the advanced ML-based ones. This has created a knowledge gap between the operators and algorithm developers. Many MLC are proprietary and even when accessible, require extensive expertise for validation. This knowledge barrier often translates to a trust gap, prompting operators to opt for rule-based controls for safety reasons.

Our framework aims to bridge this gap by providing building operators with a tool that not only offers detailed documentation but also interactive Q&A capabilities to enhance their understanding of ML-based controls. The framework enables operators to engage with the tool in the following ways:

- Adaptation Period: Initially, operators might engage more frequently with the tool, querying it to understand the rationale behind specific ML decisions. This interaction is crucial during the first few weeks post-implementation, as it helps operators gain confidence in the tool's recommendations.
- Daily Operations: During daily operations, building operators can consult the tool for hourly performance reports and recommendations. This daily interaction helps operators understand the ML-based decisions and apply them effectively, ensuring that the building's energy management aligns with its unique requirements and constraints.
- Long-term Use and Troubleshooting: Over time, the frequency of interactions may decrease, becoming more focused on occasions when inaccuracies or anomalies arise. The framework's capability to facilitate detailed inquiries allows operators to quickly identify and address issues with "bad models," which might compromise the efficiency, safety, or comfort of building occupants.

# 5 Conclusion

This study introduced an innovative framework that combines the principles of Shapley values and the in-context learning feature of LLMs to enhance the interpretability of MLC in buildings. While the Shapley values are instrumental in dissecting the contributions of various features in the ML models, LLM provides an in-depth understanding of the non-data-driven or rule-based elements in MLC. Combining them, LLM further packages these insights into a coherent, human-understandable narrative. This dual-stage interpretability ensures a robust understanding of the control decisions made. Besides directly resorting to rule-based explanations, our approach also leverages LLM to create a responsive system capable of addressing specific queries from building operators, further enhancing trust and understanding. Through the illustrative case study, the framework has showcased its potential in bridging the trust gap between algorithm developers and building operators, a crucial step in ensuring the successful adoption of advanced MLC in real-world settings.

The introduction of this framework could redefine the approach to MLC development for buildings and influence how it is utilized and trusted in building energy management. This method has the potential to detect unusual and counter-intuitive predictions, playing a critical role in 1) identifying potential issues in MLC that could lead to inefficiencies or hazardous situations, and 2) safeguarding the building infrastructure against cyber threats.

However, the framework depends significantly on domain expertise to design the prompt templates. This manual step can introduce biases and might not be scalable for larger, more complex scenarios. Advancing control-specialized LLM agents for comprehensive control tasks remains an area for future exploration. Moreover, another limitation of the study is that it is a simulation-based study with simplification of HVAC schedules and operations; future research can enhance this work by applying it in an actual building environment. We acknowledge the limitations of our validation process, primarily constrained by time and the resources required for a more extensive examination and more diverse questions to be tested.